\documentclass{article}
\usepackage[utf8]{inputenc}
\usepackage{times}
\usepackage{graphicx} 
\usepackage{subfigure} 
\usepackage{bm}
\usepackage[sort]{natbib}

\usepackage{algorithm}
\usepackage{algorithmic}

\usepackage{hyperref}

\usepackage[accepted]{icml2014} 

\newcommand{\xx}{{\bf x}}

\newcommand{\yy}{{\bf y}}
\newcommand{\vv}{{\bf v}}
\newcommand{\hh}{{\bf h}}

\newcommand{\bfl}{{\bf l}}
\newcommand{\WW}{{\bf W}}
\newcommand{\VV}{{\bf V}}

\newcommand{\bb}{{\bf b}}
\newcommand{\cc}{{\bf c}}
\newcommand{\bphi}{{\bm \phi}}
\newcommand{\ppi}{{\bm \pi}}

\newcommand{\sigm}{\mathrm{sigm}}

\icmltitlerunning{An Autoencoder Approach to Learning Bilingual Word Representations}

\begin{document} 

\twocolumn[
\icmltitle{An Autoencoder Approach to Learning Bilingual Word Representations}

\icmlauthor{Sarath Chandar A P $~^*$}{apsarathchandar@gmail.com}
\icmladdress{Indian Institute of Technology Madras, India.}
\icmlauthor{Stanislas Lauly$~^*$}{stanislas.lauly@usherbrooke.ca}
\icmladdress{Universit\'e de Sherbrooke, Canada.}
\icmlauthor{Hugo Larochelle}{hugo.larochelle@usherbrooke.ca}
\icmladdress{Universit\'e de Sherbrooke, Canada.}
\icmlauthor{Mitesh M. Khapra}{mikhapra@in.ibm.com}
\icmladdress{IBM Research India.}
\icmlauthor{Balaraman Ravindran}{ravi@cse.iitm.ac.in}
\icmladdress{Indian Institute of Technology Madras, India.}
\icmlauthor{Vikas Raykar}{viraykar@in.ibm.com}
\icmladdress{IBM Research India.}
\icmlauthor{Amrita Saha}{amrsaha4@in.ibm.com}
\icmladdress{IBM Research India.}

\icmlkeywords{boring formatting information, machine learning, ICML}

$~^*$ Both authors contributed equally.

\vskip 0.3in
]

\begin{abstract} 
Cross-language learning allows us to use training data from one language to build
models for a different language. Many approaches to bilingual learning require that we
have word-level alignment of sentences from parallel corpora. In this work we explore
the use of autoencoder-based methods for cross-language learning of vectorial word representations 
that are aligned between two languages, while not relying on word-level alignments. 
We show that by simply learning to reconstruct the bag-of-words representations of aligned sentences,
within and between languages, we can in fact learn high-quality representations and do without word alignments. 
Since training autoencoders on word
observations presents certain computational issues, we propose and compare
different variations adapted to this setting.
We also propose an explicit correlation maximizing regularizer that leads to significant
improvement in the performance. We empirically investigate the success of our approach
on the problem of cross-language
test classification, where a classifier trained on a given language
(e.g.,\ English) must learn to generalize to a different language (e.g.,\ German). 
These experiments demonstrate that our approaches are competitive
with the state-of-the-art, achieving up to 10-14 percentage point 
improvements over the best reported results on this task.
\end{abstract}

\section{Introduction}
Languages such as English, which have plenty of annotated resources at their disposal have better Natural Language Processing (NLP) capabilities than other languages that are not so fortunate in terms of annotated resources. For example, high quality POS taggers \cite{Toutanova:2003:FPT:1073445.1073478}, parsers \cite{socher-EtAl:2013:ACL2013}, sentiment analyzers \cite{DBLP:series/synthesis/2012Liu} are already available for English but this is not the case for many other languages such as Hindi, Marathi, Bodo, Farsi, Urdu, \textit{etc}. This situation was acceptable in the past when only a few languages dominated the digital content available online and elsewhere. However, the ever increasing number of languages on the web today has made it important to accurately process natural language data in such lesser-fortunate languages also. An obvious solution to this problem is to improve the annotated inventory of these languages but the involved cost, time and effort act as a natural deterrent to this.    

Another option is to exploit the unlabeled data available in a language. In this context, vectorial text representations have proven useful for multiple NLP
tasks~\citep{Turian+Ratinov+Bengio-2010,CollobertR2011}. It's been shown that meaningful representations, capturing
syntactic and semantic similarity, can be learned from unlabeled
data. Along with a (usually smaller) set of labeled data, these
representations allow to exploit unlabeled data and improve
the generalization performance on some given task, even allowing to
generalize out of the vocabulary observed in the labeled data only (thereby, partly alleviating the problem of data sparsity).

While the majority of previous work on vectorial text representations has concentrated on the
monolingual case, recent work has started looking at learning word
and document representations that are aligned across languages ~\citep{KlementievA2012,ZhouW2013,MikolovT2013}. Such aligned representations can potentially allow the use of resources from a resource fortunate language to develop NLP capabilities in a resource deprived language ~\citep{a1,a2,a3,a4,a5}.
For example, if a common representation model is learned for representing English and German documents, then a classifier trained on annotated English documents can be used to classify German documents (provided we use the learned common representation model for representing documents in both languages). 

Such reuse of  resources across languages has been tried in the past by projecting parameters learned from the annotated data of one language to another language ~\cite{a1,a2,a3,a4,a5}
These projections are enabled by a bilingual resource such as a Machine Translation (MT) system. Recent attempts at learning common bilingual representations ~\cite{KlementievA2012,ZhouW2013,MikolovT2013} aim to eliminate the need of such a MT system. Such bilingual 
representations have been applied to a variety of problems, including
cross-language document classification~\cite{KlementievA2012} and phrase-based
machine translation~\cite{ZhouW2013}. A common property of these approaches
is that a {\it word-level alignment} of translated sentences is leveraged,
\textit{e.g.},\ to derive a regularization term relating word embeddings
across languages~\cite{KlementievA2012,ZhouW2013}. Such methods not only eliminate the need for an MT system but also outperform MT based projection approaches. 

In this paper, we experiment with a method to learn
bilingual word representations that {\it does without word-to-word
alignment} of bilingual corpora during training. Unlike previous approaches ~\cite{KlementievA2012}, we only require
aligned sentences and do not rely on word-level alignments
(e.g.,\ extracted using GIZA++, as is usual), which simplifies the learning procedure. To do so, we propose a
bilingual autoencoder model, that learns hidden
encoder representations of paired bag-of-words sentences which are not only informative of the original bag-of-words
but also predictive of each other. Word representations can then easily be extracted from the encoder
and used in the context of a supervised NLP task. Specifically, we demonstrate the quality of these representations for
the task of cross-language document
classification, where a labeled data set can be available in one language, but not in another one.
As we'll see, our approach is able to reach state-of-the-art performance, achieving up to 10-14 percentage point 
improvements over the best previously reported results.


\section{Autoencoder for Bags-of-Words}
\label{sec:autoencoder}

Let $\xx$ be the bag-of-words representation of a
sentence. Specifically, each $x_i$ is a word index from a fixed
vocabulary of $V$ words. As this is a bag-of-words, the order of the
words within $\xx$ does not correspond to the word order in the
original sentence. We wish to learn a $D$-dimensional vectorial
representation of our words from a training set of sentence
bag-of-words $\{\xx^{(t)}\}_{t=1}^T$.

We propose to achieve this by using an autoencoder model that encodes
an input bag-of-words $\xx$ with a sum of the representations
(embeddings) of the words present in $\xx$, followed by a nonlinearity. Specifically, let matrix $\WW$ be the $D\times V$ 
matrix whose columns are the
vector representations for each word. The encoder's computation will involve summing over the columns of $\WW$ for each
word in the bag-of-word. We will note this encoder function $\bphi(\xx)$.
Then, using a decoder, the autoencoder will be trained to optimize
a loss function that measures how predictive of the original bag-of-words the 
 encoder representation $\bphi(\xx)$ is.

There are different variations we can consider, in the design of the encoder/decoder and the choice of loss function. One must be careful however, as certain choices can be inappropriate for training on word observations, which
are intrinsically sparse and high-dimensional. In this paper, we explore and compare two different approaches, described in the next two sub-sections.


\subsection{Binary bag-of-words reconstruction training with merged mini-batches}
\label{sec:reconstruction_sampling}
In the first approach, we start from the conventional autoencoder architecture, which minimizes a cross-entropy loss that compares a binary vector observation with a decoder reconstruction. We thus convert the bag-of-words $\xx$ into a fixed-size but sparse binary vector $\vv(\xx)$, which is such that $v(\xx)_{x_i}$ is 1 if word $x_i$ is present in $\xx$ or otherwise 0.

From this representation, we obtain an encoder representation by multiplying $\vv(\xx)$ with the word representation matrix $\WW$
\begin{equation}
\bphi(\xx) = \hh(\cc + \WW \vv(\xx))
\end{equation}
where $\hh(\cdot)$ is an element-wise non-linearity such as the
sigmoid or hyperbolic tangent, and $\cc$ is a
$D$-dimensional bias vector. Encoding thus involves summing the word representation of the words present at least once in the bag-of-word.

To produce a reconstruction, we parametrize the decoder using the following non-linear form:
\begin{equation}
  \widehat{\vv}(\xx) = \sigm(\VV\bphi(\xx) + \bb)
\end{equation}
where $\VV$ = $\WW^T$ and $\bb$ is the bias vector of the reconstruction layer and $\sigm(a) = 1 /
(1+\exp(-a))$ is the sigmoid non-linearity.
  
Then, the reconstruction
is compared to the original binary bag-of-words as follows:
\begin{equation}
\ell(\vv(\xx)) = - \sum_{i=1}^{V} v(\xx)_i \log( \widehat{v}(\xx)_i) + (1 - v(\xx)_i) \log (1 - \widehat{v}(\xx)_i)~.
\end{equation} 
Training then proceeds by optimizing the sum of reconstruction cross-entropies across the training set, \textit{e.g.},\
using stochastic or mini-batch gradient descent.

Note that, since the binary bag-of-words are very high-dimensional (the dimensionality corresponds to the
size of the vocabulary, which is typically large), the above training procedure which aims at reconstructing the complete binary bag-of-word, will be slow. Since we will later be training on millions of sentences, training on each individual sentence bag-of-words will be expensive.

Thus, we propose a simple trick, which exploits the bag-of-words structure of the input. Assuming we are performing mini-batch training (where a mini-batch contains a list of bag-of-words), we simply propose to merge the bags-of-words of the mini-batch into a single bag-of-word, and revert back to stochastic gradient descent. The resulting effect is that each update is as efficient as in stochastic gradient descent, but the number of updates per training epoch is divided by the mini-batch size. As we'll see in the experimental section, we've found this trick to still produces good word representations, while sufficiently reducing training time.

We note that, additionally, we could have used the stochastic approach proposed by \citet{DauphinY2011} for reconstructing binary bag-of-words representations of documents, to further improve the efficiency of training. They use importance sampling to avoid reconstructing the whole $V$-dimensional input vector.

\subsection{Tree-based decoder training}
\label{sec:tree-based_training}

The previous autoencoder architecture worked with a binary vectorial
representation of the input bag-of-word. In the
second autoencoder architecture we investigated, we considered
an architecture that instead works with the bag (unordered list) representation
more directly. 

Firstly, the encoder representation will now involve a sum of the representation of all words, reflecting the relative frequency of each word:
\begin{equation}
\bphi(\xx) = \hh\left(\cc + \sum_{i=1}^{|\xx|} \WW_{\cdot,x_i}\right)~.
\end{equation}
Notice that this implies that the scaling of the pre-activation (the input 
to the nonlinearity) can vary between bags-of-words, depending on how many
words it contains. Thus, we'll optionally consider using the average of the representations, as opposed to the sum (this choice is cross-validated in our experiments).

Moreover, decoder training will assume that, from
the decoder's output, we can obtain a probability distribution over any word
$\widehat{x}$ observed at the reconstruction output layer
$p(\widehat{x}|\bphi(\xx))$. Then, we can treat the input bag-of-words as
a $|\xx|$-trials multinomial sample from that distribution and use as
the reconstruction loss its negative log-likelihood:
\begin{equation}
  \ell(\xx) = \sum_{i=1}^{V} -\log p(\widehat{x}=x_i|\bphi(\xx))~.
\end{equation}
We now must ensure that the decoder can compute $p(\widehat{x}=x_i|\bphi(\xx))$
efficiently from $\bphi(\xx)$. Specifically, we'd like to avoid a
procedure scaling linearly with the vocabulary size $V$, since $V$ will
be very large in practice. This precludes any procedure that would
compute the numerator of $p(\widehat{x}=w|\bphi(\xx))$ for each possible word $w$
separately and normalize so it sums to one.

We instead opt for an approach borrowed from the work on neural
network language models~\cite{Morin+al-2005,Mnih+Hinton-2009}. Specifically, we use a probabilistic
tree decomposition of $p(\widehat{x}=x_i|\bphi(\xx))$.  Let's assume each
word has been placed at the leaf of a binary tree.  We can then treat
the sampling of a word as a stochastic path from the root of the tree
to one of the leaves. 

We denote as $\bfl(x)$ the sequence of
internal nodes in the path from the root to a given word $x$, with
$l(x)_1$ always corresponding to the root. We will denote as
$\ppi(x)$ the vector of associated left/right branching choices on
that path, where $\pi(x)_k=0$ means the path branches left at internal
node $l(x)_k$ and branches right if $\pi(x)_k=1$ otherwise.  Then, the
probability $p(\widehat{x}=x|\bphi(\xx))$ of reconstructing a certain word $x$ observed in the
bag-of-words is computed as
\begin{equation}
  p(\widehat{x}|\bphi(\xx)) = \prod_{k=1}^{|\ppi(\hat{x})|} p(\pi(\widehat{x})_k|\bphi(\xx))
\end{equation}
where $p(\pi(\widehat{x})_k|\bphi(\xx))$ is output by the decoder. By using a full
binary tree of words, the number of different decoder outputs required
to compute $p(\widehat{x}|\bphi(\xx))$ will be logarithmic in the vocabulary size
$V$. Since there are $|\xx|$ words in the bag-of-words, at most
$O(|\xx| \log V)$ outputs are required from the decoder. This is
of course a worst case scenario, since words will share internal nodes
between their paths, for which the decoder output can be computed just
once. As for organizing words into a tree, as in \citet{LarochelleH2012} we used a
random assignment of words to the leaves of the full binary tree,
which we have found to work well in practice.

Finally, we need to choose a parametrized form for the decoder.
We choose the following non-linear form:
\begin{equation}
  p(\pi(\widehat{x})_k=1|\bphi(\xx)) = \sigm(b_{l(\hat{x}_i)_k} + \VV_{l(\hat{x}_i)_k,\cdot} \bphi(\xx))
\end{equation}
where $\bb$ is a ($V$-1)-dimensional bias
vector and $\VV$ is a $(V-1)\times D$ matrix. Each left/right
branching probability is thus modeled with a logistic regression model
applied on the encoder representation of the input
bag-of-words $\bphi(\xx)$.

\section{Bilingual autoencoders}
\label{sec:bilingual_autoencoder}

Let's now assume that for each sentence bag-of-words $\xx$ in some
source language ${\cal X}$, we have an associated bag-of-words $\yy$ for the same
sentence translated in some target language ${\cal Y}$ by a human expert. 

Assuming we have a training set of such $(\xx,\yy)$ pairs, we'd like
to use it to learn representations in both languages that are aligned,
such that pairs of translated words have similar representations.

To achieve this, we propose to augment the regular autoencoder
proposed in Section~\ref{sec:autoencoder} so that, from the sentence
representation in a given language, a reconstruction can be attempted
of the original sentence in the other language. Specifically, we now define language specific word representation
matrices $\WW^{x}$ and $\WW^{y}$, corresponding to the languages
of the words in $\xx$ and $\yy$ respectively. Let $V^{\cal X}$ and 
$V^{\cal Y}$ also be the number of words in the vocabulary of both languages,
which can be different. The word representations however are of
the same size $D$ in both languages. For the binary reconstruction
autoencoder, the bag-of-words representations
extracted by the encoder becomes
\begin{eqnarray*}
  \bphi(\xx) = \hh\left(\cc + \WW^{\cal X} \vv(\xx)\right)~,~~ \bphi(\yy) = \hh\left(\cc + \WW^{\cal Y} \vv(\yy)\right)
\end{eqnarray*}
and are similarly extended for the tree-based autoencoder.
Notice that we share the
bias $\cc$ before the nonlinearity across encoders,
to encourage the encoders in both languages to produce representations
on the same scale.

\begin{figure}[t]
\vskip 0.2in
\begin{center}
\centerline{\includegraphics[width=0.4\textwidth]{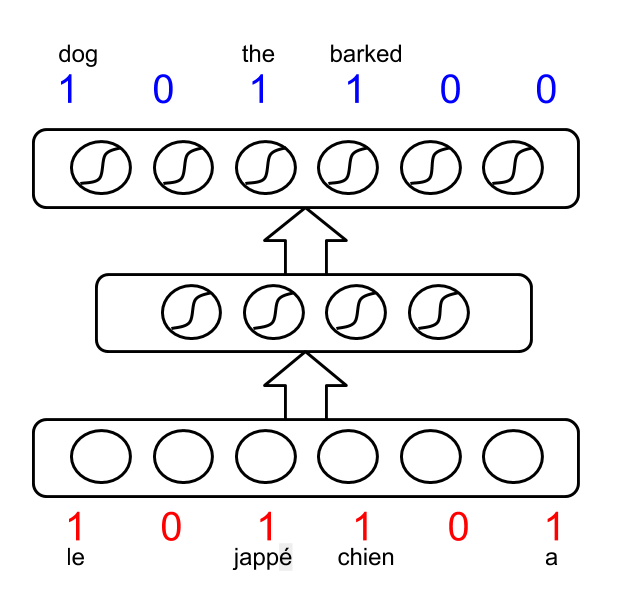}}
\caption{Illustration of a binary reconstruction error based bilingual autoencoder that learns to reconstruct
the binary bag-of-words of the English sentence ``{\it the dog barked}''
from its French translation ``{\it le chien a jappé}''.}
\label{fig:errorbased_bilingual_autoencoder}
\end{center}
\vskip -0.2in
\end{figure}

\begin{figure}[t]
\begin{center}
\includegraphics[width=0.4\textwidth]{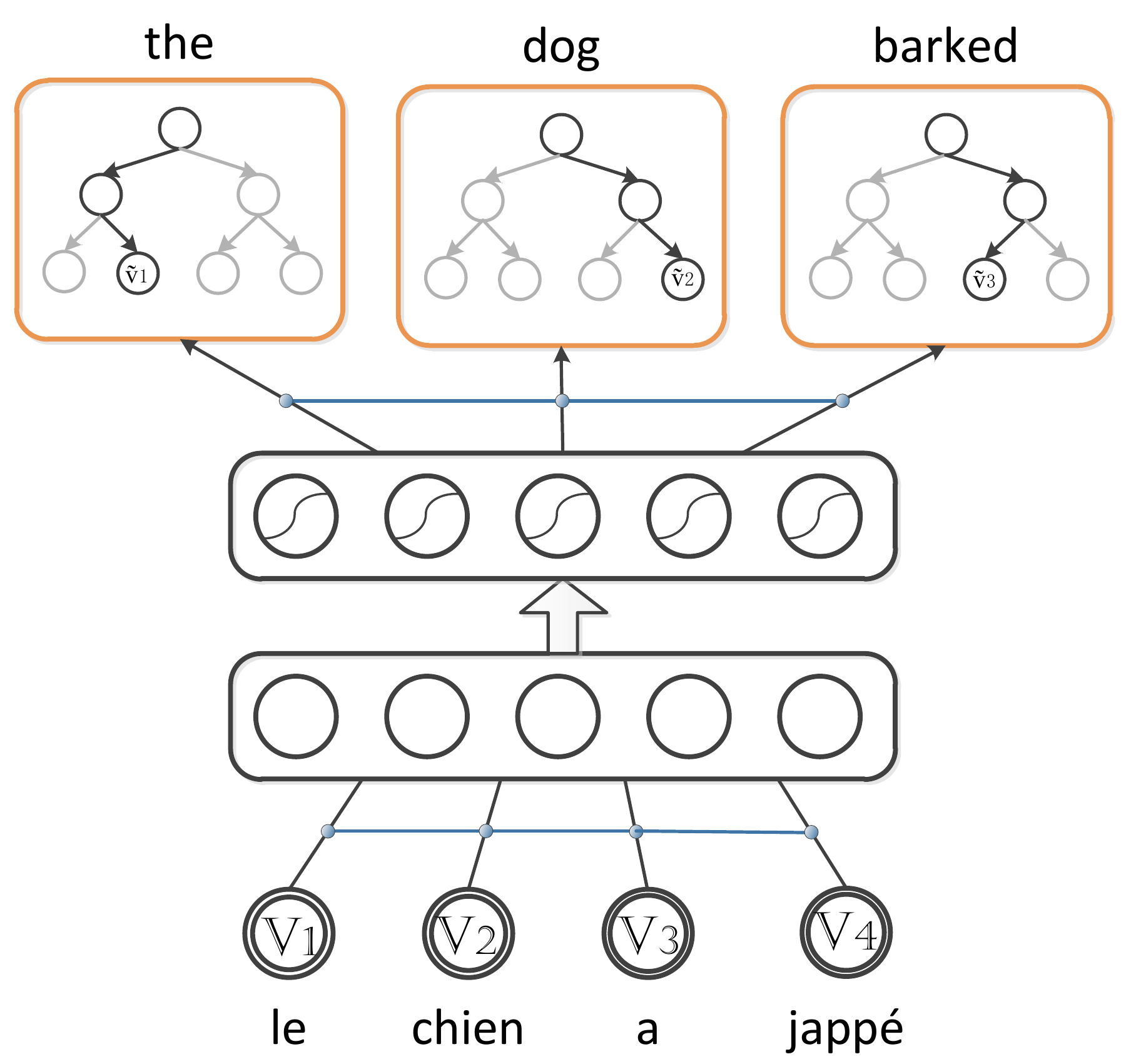}
\end{center}
\caption{Illustration of a tree-based bilingual autoencoder that learns to construct
the bag-of-words of the English sentence ``{\it the dog barked}''
from its French translation ``{\it le chien a jappé}''. The
horizontal blue line across the input-to-hidden connections
highlights the fact that these connections share the same parameters
(similarly for the hidden-to-output connections).}
\label{fig:treebased_bilingual_autoencoder}
\end{figure}

From the sentence in either languages, we want to be able to perform a
reconstruction of the original sentence in any of the languages. In
particular, given a representation in any language, we'd like a
decoder that can perform a reconstruction in language ${\cal X}$ and
another decoder that can reconstruct in language ${\cal Y}$.  Again,
we use decoders of the form proposed in either Section~\ref{sec:reconstruction_sampling}~or~\ref{sec:tree-based_training} (see Figures \ref{fig:errorbased_bilingual_autoencoder} and \ref{fig:treebased_bilingual_autoencoder}),
but let the decoders of each language have their own parameters
$(\bb^{\cal X},\VV^{\cal X})$ and $(\bb^{\cal Y},\VV^{\cal Y})$.

This encoder/decoder decomposition structure allows us to learn a mapping within
each language and across the languages. Specifically, for a given pair
$(\xx,\yy)$, we can train the model to (1) construct $\yy$ from $\xx$ (loss $\ell(\xx,\yy)$),
(2) construct $\xx$ from $\yy$ (loss $\ell(\yy,\xx)$), (3) reconstruct $\xx$ from itself (loss $\ell(\xx)$) and
(4) reconstruct $\yy$ from itself (loss $\ell(\yy)$). We follow this approach
in our experiments and optimize the sum of the corresponding 4 losses during training.


\subsection{Cross-lingual correlation regularization}

\label{sec:correlation}
The bilingual encoder proposed above can be further enriched by ensuring that the embeddings learned for a given pair $(\xx,\yy)$ are highly correlated. We achieve this by adding a correlation term to the objective function. Specifically, we could optimize
\begin{equation}
  \ell(\xx,\yy) +\ell(\yy,\xx)  - \lambda \cdot cor(\bphi(\xx), \bphi(\yy))
\end{equation}
where $cor(\bphi(\xx), \bphi(\yy))$ is the correlation between the encoder representations learned for $\xx$ and $\yy$ and $\lambda$ is a scaling factor which ensures that the three terms in the loss function have the same range. Note that this approach could be used for either the binary bag-of-words or the tree-based reconstruction autoencoders.

\subsection{Document representations}
\label{sec:docrep}
Once we learn the language specific word representation
matrices $\WW^{x}$ and $\WW^{y}$ as described above, we can use them to construct document representations, by using their columns as vector representations for words in both languages. Now, given a document ${\bf d}$ written in language ${\cal Z}\in \{{\cal X},{\cal Y}\}$ and containing $m$ words, $z_1, z_2, \dots, z_m$, we represent it as the tf-idf weighted sum of its words' representations:

\begin{equation}
  \psi({\bf d}) = \sum_{i=1}^{m} \textit{tf-idf}(z_i)\cdot\WW_{.,z_i}^{\cal Z}
\end{equation}

We use the document representations thus obtained to train our document classifiers, in the cross-lingual document classification task described in Section~\ref{exp}.

\section{Related Work}
\label{sec:related_work}


Recent work that has considered the problem of learning bilingual representations of words usually has relied on word-level
alignments. \citet{KlementievA2012} propose to train simultaneously two neural
network languages models, along with a regularization term that
encourages pairs of frequently aligned words to have similar word
embeddings.  \citet{ZhouW2013} use a similar approach, with a different form
for the regularizer and neural network language models as
in~\cite{CollobertR2011}. In our work, we specifically investigate whether
a method that does not rely on word-level alignments can
learn comparably useful multilingual embeddings in the
context of document classification.

Looking more generally at neural networks that learn multilingual
representations of words or phrases, we mention the work of \citet{GaoJ2013}
which showed that a useful linear mapping between {\it separately
  trained} monolingual skip-gram language models could be learned. 
They too however rely on the specification of pairs of words
in the two languages to align. \citet{MikolovT2013} also propose a method
for training a neural network to learn useful representations
of phrases (i.e.\ short segments of words), in the context of
a phrase-based translation model. In this case, phrase-level
alignments (usually extracted from word-level alignments)
are required.

\section{Experiments}
\label{exp}

The techniques proposed in this paper enable us to learn bilingual embeddings which capture cross-language similarity between words. We propose to evaluate the quality of these embeddings by using them for the task of cross-language document classification. We follow the same setup as used by \citet{KlementievA2012} and compare with their method. The set up is as follows. A labeled data set of documents in some language ${\cal X}$ is available to train a classifier, however we are interested in classifying documents in a different language ${\cal Y}$ at test time. To achieve this, we leverage some bilingual corpora, which importantly is not labeled with any document-level categories. This bilingual corpora is used instead to learn document representations in both languages ${\cal X}$ and ${\cal Y}$ that are encouraged to be invariant to translations from one language to another. The hope is thus that we can successfully apply the classifier trained on document representations for language ${\cal X}$ directly to the document representations for language ${\cal Y}$. We use English (EN) and German (DE) as the language pair for all our experiments. 

\subsection{Data}
For learning the bilingual embeddings, we used the English German section of the Europarl corpus \cite{europ} which contains roughly 2 million parallel sentences. As mentioned earlier, unlike \citet{KlementievA2012}, we do not use any word alignments between these parallel sentences. We use the same pre-processing as used by \citet{KlementievA2012}. Specifically, we tokenize the sentences using NLTK \cite{nltk}, remove punctuations and lowercase all words. We do not remove stopwords (similar to ~\citet{KlementievA2012}).   

Note that \citet{KlementievA2012} use the word-aligned Europarl corpus to first learn an interaction matrix between the words in the two languages. This interaction matrix is then used in a multitask learning setup to induce bilingual embeddings from English and German sections of the Reuters RCV1/RCV2 corpora. Note that these documents are not parallel. Each document is assigned one or more categories from a pre-defined hierarchy of topics. Following \citet{KlementievA2012}, we consider only those documents which were assigned exactly one of the 4 top level categories in the topic hierarchy. These topics are CCAT (Corporate/Industrial), ECAT (Economics), GCAT (Government/Social) and MCAT (Markets). The number of such documents sampled by \citet{KlementievA2012} for English and German is 34,000 and 42,753 respectively. In contrast to \citet{KlementievA2012}, we do not require a two stage approach (of learning an interaction matrix and then inducing bilingual embeddings). We directly learn the embeddings from the Europarl corpus which is parallel. Further, in addition to the Europarl corpus, we also considered feeding the same RCV1/RCV2 documents (34000 EN and 42,753 DE) to the autoencoders. These non-parallel documents are used only to reinforce the monolingual embeddings (by reconstructing $\xx$ from $\xx$ or $\yy$ from $\yy$). So, in effect, we use the same amount of data as that used by \citet{KlementievA2012} but our model/training procedure is completely different.

Next for the cross language classification experiments, we again follow the same setup as used by \citet{KlementievA2012}. Specifically, we use 10,000 single-topic documents for training and 5000 single-topic documents for testing in each language. These documents are also pre-processed using a similar procedure as that used for the Europarl corpus. 

\begin{table*}[!t]
\begin{center}
\caption{Example English words along with 10 closest words both in English (en) and German (de), using the Euclidean distance between the embeddings learned by BAE-cr/corr}
\begin{tabular}{cc|cc|cc}
  
   \multicolumn{2}{c|}{january} & \multicolumn{2}{|c|}{president} & \multicolumn{2}{|c}{said}\\ 
  \hline 
  en & de & en & de & en & de \\
  \hline
  january&januar&president&präsident&said&gesagt\\
march&märz&i&präsidentin&told&sagte\\
october&oktober&mr&präsidenten&say&sehr\\
july&juli&presidents&herr&believe&heute\\
december&dezember&thank&ich&saying&sagen\\
1999&jahres&president-in-office&ratspräsident&wish&heutigen\\
june&juni&report&danken&shall&letzte\\
month&1999&voted&danke&again&hier\\
year&jahr&colleagues&bericht&agree&sagten\\
september&jahresende&ladies&kollegen&very&will\\
\hline
  \hline

   \multicolumn{2}{c|}{oil} & \multicolumn{2}{|c|}{microsoft} & \multicolumn{2}{|c}{market}\\ 
  \hline 
  en & de & en & de & en & de \\
  \hline
  oil&öl&microsoft&microsoft&market&markt\\
supply&boden&cds&cds&markets&marktes\\
supplies&befindet&insider&warner&single&märkte\\
gas&gerät&ibm&tageszeitungen&commercial&binnenmarkt\\
fuel&erdöl&acquisitions&ibm&competition&märkten\\
mineral&infolge&shareholding&handelskammer&competitive&handel\\
petroleum&abhängig&warner&exchange&business&öffnung\\
crude&folge&online&veranstalter&goods&binnenmarktes\\
materials&ganze&shareholder&geschäftsführer&sector&bereich\\
causing&nahe&otc&aktiengesellschaften&model&gleichzeitig\\
  \hline

  \end{tabular}
  
  \label{Tabd}   
  \end{center}
\end{table*}

\subsection{Cross language classification}
Our overall procedure for cross language classification can be summarized as follows:

\begin{itemize}
\item Train bilingual word representations $\WW^{x}$ and $\WW^{y}$ on sentence pairs extracted from Europarl-v7 for languages ${\cal X}$ and ${\cal Y}$. Optionally, we also use the monolingual documents from RCV1/RCV2 to reinforce the monolingual embeddings (this choice is cross-validated, as described in Section~\ref{sec:emb-models}). 
\item Train document classifier on the Reuters training set for language ${\cal X}$, where documents are represented using the word representations $\WW^{x}$.
\item Use the classifier trained in the previous step on the Reuters test set for language ${\cal Y}$, using the word representations $\WW^{y}$ to represent the documents.
\end{itemize}

As in \citet{KlementievA2012} we used an averaged perceptron to train a multi-class classifier for 10 epochs, for all the experiments (\citet{KlementievA2012} report that the results were not sensitive to the number of epochs). The English and German vocabularies contained 43,614 and 50,110 words, respectively. Each document is represented with the tf-idf weighted linear combination of its word's embeddings, as described in Section~\ref{sec:docrep}, where only the words belonging to the above vocabulary are considered.

\subsection{Different models for learning embeddings}
\label{sec:emb-models}
From the different autoencoder architectures and the optional correlation-based regularization term proposed earlier, we trained 3 different models for learning bilingual embeddings. Each of these models is described below. 

\begin{itemize}
\item BAE-tr: uses tree-based decoder training (see Section \ref{sec:tree-based_training}). 
\item BAE-cr: uses reconstruction error based decoder training (see Section \ref{sec:reconstruction_sampling}). 
\item BAE-cr/corr: uses reconstruction error based decoder training (see Section \ref{sec:reconstruction_sampling}), but unlike BAE-cr it uses the correlation based regularization term (see Section \ref{sec:correlation}). 
\end{itemize}
As we'll see, BAE-cr is our worse performing model, thus this experiment will allow us to observe whether the correlation regularization can play an important role in improving the quality of the representations.

All of the above models were trained for up to 20 epochs using the same data as described earlier. All results are for word embeddings of size $D=40$, as in \citet{KlementievA2012}. Further, to speed up the training for BAE-cr and BAE-cr/corr we merged mini-batches of 5 adjacent sentence pairs into a single training instance, as described in Section~\ref{sec:reconstruction_sampling}. 

Other hyperparameters were selected using a training/validation set split of 80\% and 20\% and using the performance on the validation set of an averaged perceptron trained on the smaller training set portion (notice that this corresponds to a monolingual classification experiment, since the general assumption is that no labeled data is available in the test set language).

\section{Results and Discussions}
Before discussing the results of cross language classification, we would first like to give a qualitative feel for the embeddings learned by our method. For this, we perform a small experiment where we select a few English words and list the top 10 English and German words which are most similar to these words (in terms of the Euclidean distance between their embeddings as learned by BAE-cr/corr). Table \ref{Tabd} shows the result of this experiment. For example, Table \ref{Tabd} shows that in all the cases the German word which is closest to a given English word is actually the translation of that English word. Also, notice that the model is able to capture semantic similarity between words by embedding semantically similar words (such as, \textit{(january, march)}, \textit{(gesagt, sagte)}, \textit{(market, commercial)}, \textit{etc.}) close to each other. The results of this experiment suggest that these bilingual embeddings should be useful for any cross language classification task as indeed shown by the results presented in the next section. The supplementary material also includes a 2D visualization of the word embeddings in both languages, generated using the t-SNE dimensionality reduction algorithm~\citep{VanDerMaaten08}.

\subsection{Comparison of the performance of different models}

We now present the cross language classification results obtained by using the embeddings produced by each of the 3 models described above. We also compare our models with the following approaches, for which the results are reported in \citet{KlementievA2012}:

\begin{itemize}

\item Klementiev et al. : This model uses word embeddings learned by a multitask neural network language model with a regularization term that
encourages pairs of frequently aligned words to have similar word
embeddings. From these embeddings, document representations are computed as described in Section~\ref{sec:docrep}. 

\item MT: Here, test documents are translated to the language of the training documents using a Machine Translation (MT) system. MOSES\footnote{\url{http://www.statmt.org/moses/}}, a standard phrase-based MT system, using default parameters and a 5-gram language model was trained on the Europarl v7 corpus (same as the one used for inducing our bilingual embeddings).

\item Majority Class: Every test document is simply assigned the Majority class prevalent in the training data.

\end{itemize}

Table \ref{tab:title} summarizes the results obtained using 1K training data with different models. We report results in both directions, \textit{i.e.}, EN-DE and DE-EN. Between BAE-tr and BAE-cr, we observe that BAE-tr provides better performance and is comparable to the embeddings learned by the neural network language model of \citet{KlementievA2012} which, unlike BAE-tr, relies on word-level alignments. We also observe that the use of the correlation regularization is very beneficial. Indeed, it is able to improve the performance of BAE-cr and make it the best performing method, with more than 10\% in accuracy over other methods for the EN to DE task.

\begin {table}[t]
\caption {Classification Accuracy for training on English and German with 1000 labeled examples} \label{tab:title} 
\begin{center}
\begin{tabular}{|l|l|l|}
\hline
{\bf } & EN $\rightarrow$ DE & DE $\rightarrow$ EN \\
\hline
BAE-tr & 80.2 & 68.2 \\
\hline
BAE-cr  & 78.2 & 63.6 \\
\hline
BAE-cr/corr  & {\bf 91.8} & {\bf 72.8} \\
\hline
Klementiev et al.  & 77.6 & 71.1 \\
\hline
MT  & 68.1 & 67.4 \\
\hline
Majority Class  & 46.8 & 46.8 \\
\hline
\end{tabular}
\end{center}
\end{table}

\subsection{Effect of varying training size}
\begin{figure}[t]
\begin{center}
\includegraphics[width=0.48\textwidth]{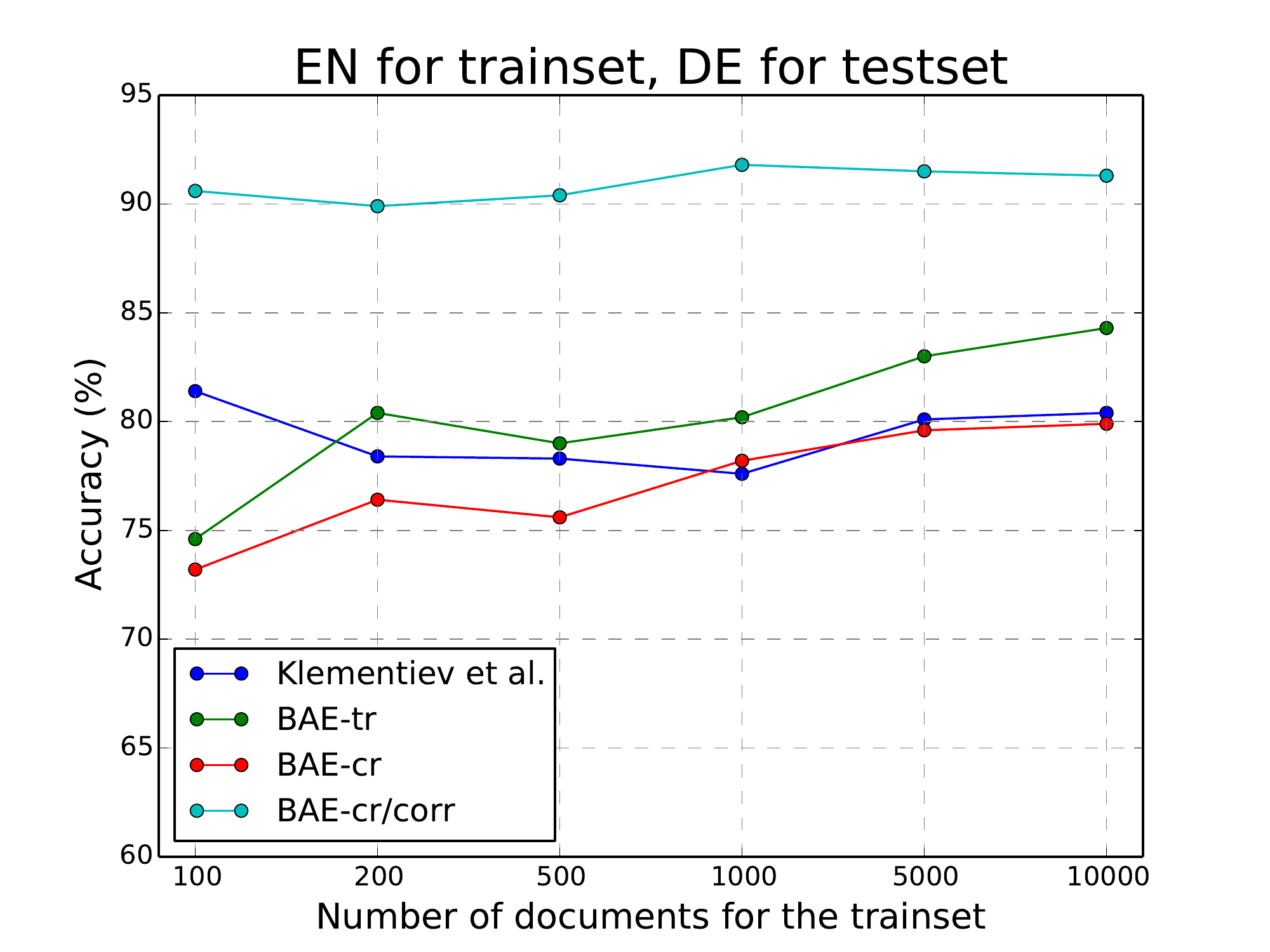}
\end{center}
\caption{Crosslingual classification accuracy results with English documents for the train set and German documents for the test set}
\label{fig:x}
\end{figure}

\begin{figure}[t]
\begin{center}
\includegraphics[width=0.48\textwidth]{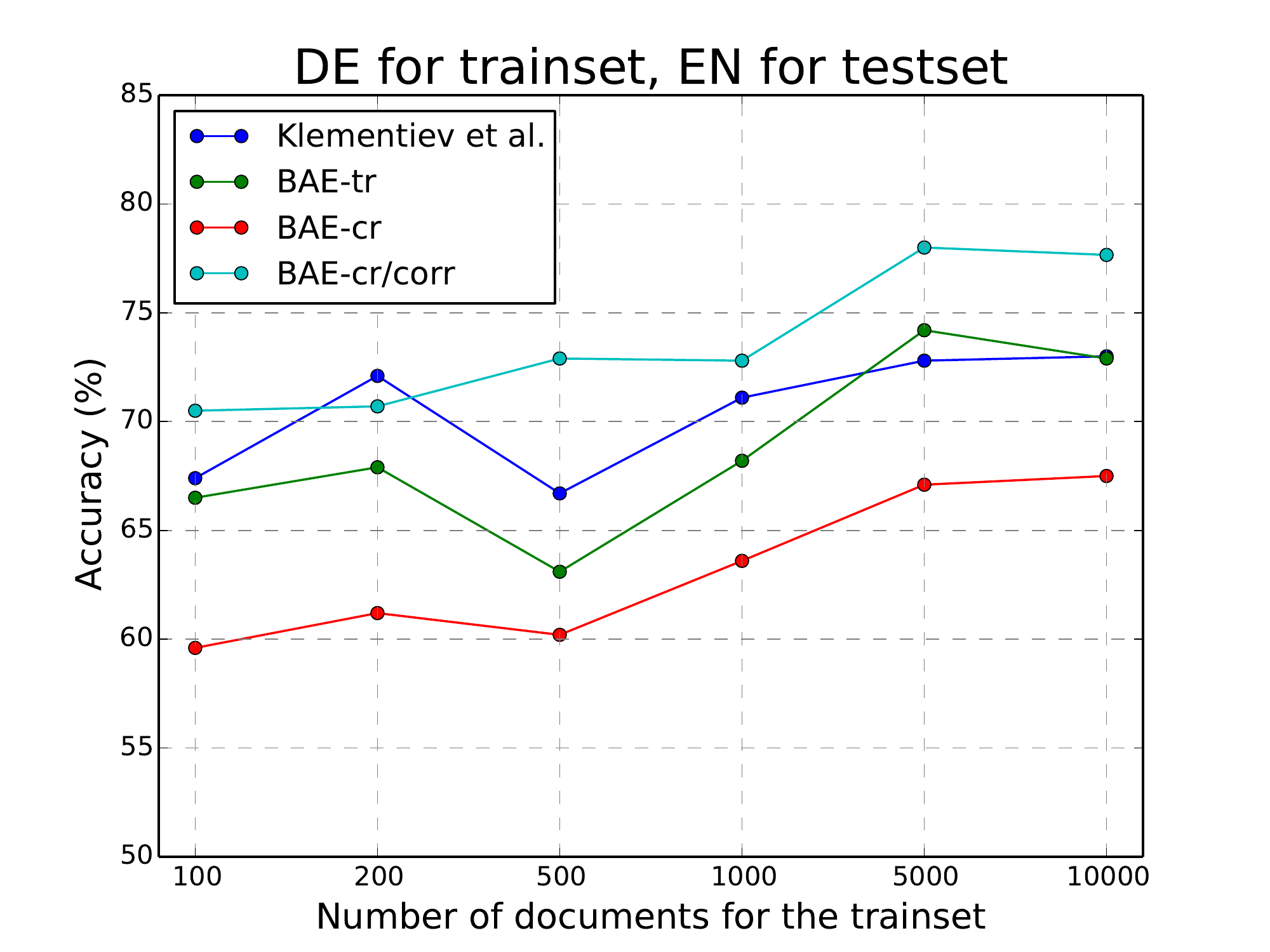}
\end{center}
\caption{Crosslingual classification accuracy results with German documents for the train set and English documents for the test set}
\label{fig:y}
\end{figure}

Next, we evaluate the effect of varying the amount of supervised training data for training the classifier, with either BAE-tr, BAE-cr/corr or \citet{KlementievA2012} embeddigns. We experiment with training sizes of 100, 200, 500, 1000, 5000 and 10000.  These results for EN-DE and DE-EN are summarized in Figure \ref{fig:x} and Figure \ref{fig:y} respectively. We observe that BAE-cr/corr clearly outperforms the other models at almost all data sizes. More importantly, it performs remarkably well at very low data sizes ($t$=100) which suggests that it indeed learns very meaningful embeddings which can generalize well even at low data sizes.

\subsection{Effect of coarser alignments}
The excellent performance of BAE-cr/corr suggests that merging mini-batches
into single bags-of-words does not significantly impact the quality of the word embeddings. In other words, not only we do not need to rely on word-level alignments, but exact sentence-level alignment is also not essential to reach good performances. It is thus natural to ask the effect of using even coarser level alignments. We check this by varying the size of the merged mini-batches from 5, 25 to 50, for both BAE-cr/corr and BAE-tr. The cross language classification results obtained by using these coarser alignments are summarized in Table~\ref{tab:title2}. 

Surprisingly, the performance of BAE-tr does not significantly decrease, by using merged mini-batches of size 5 (in fact, the performance even improves for the EN to DE task). However, with larger mini-batches, the performance can deteriorate, as is observed on the DE to EN task, for the BAE-cr/corr embeddings.

\begin {table}[t]
\caption {Classification Accuracy for training on English and German with coarser alignments for 1000 labeled examples} \label{tab:title2} 
\begin{center}
\begin{tabular}{|l|l|l|l|}
\hline
{\bf } & Sent. per doc &  EN $\rightarrow$ DE & DE $\rightarrow$ EN \\
\hline
BAE-tr & 5 & 84.0 & 67.7 \\
\hline
BAE-tr & 25 & 83.0 & 63.4 \\
\hline
BAE-tr & 50 & 75.9 & 68.6 \\
\hline

BAE-cr/corr  & 5 & 91.75 & 72.78 \\
\hline
BAE-cr/corr  & 25 & 88.0 & 64.5 \\
\hline
BAE-cr/corr  &50 & 90.2 & 49.2 \\
\hline
\end{tabular}
\end{center}
\end{table}

\section{Conclusion and Future Work}

We presented evidence that meaningful bilingual word
representations could be learned without relying on word-level
alignments and can even be successful on fairly coarse sentence-level alignments. In particular, we showed that even though our model does not use word level alignments, it is able to outperform a state of the art word representation learning method that exploits word-level alignments. In addition, it also outperforms a strong Machine Translation based baseline. We observed that using a correlation based regularization term leads to better bilingual embeddings which are highly correlated and hence perform better for cross language classification tasks.

As future work we would like to investigate extensions of our bag-of-
words bilingual autoencoder to bags-of-n-grams, where the model would also have to learn representations for short phrases. Such a model  should be particularly useful in the context of a machine translation system. We would also like to explore the possibility of converting our bilingual model to a multilingual model which can learn common representations for multiple languages given different amounts of parallel data between these languages.



\bibliographystyle{unsrtnat}
\bibliography{multilingual_autoencoder}

\end{document}